\pdfoutput=1

\documentclass[11pt]{article}

\usepackage{ACL2023}

\usepackage{times}
\usepackage{latexsym}
\usepackage[T1]{fontenc}
\usepackage[utf8]{inputenc}
\usepackage{microtype}
\usepackage{inconsolata}

\usepackage{multirow}
\usepackage{amsmath}
\usepackage{capt-of}
\usepackage{tabularx}
\usepackage{subcaption}
\usepackage{epsfig}
\usepackage{amssymb}
\usepackage{amsfonts}
\usepackage{booktabs}
\usepackage{scalerel}
\usepackage[inline]{enumitem}
\usepackage{listings}
\usepackage{varwidth}
\usepackage[export]{adjustbox}
\usepackage{tikz}
\usetikzlibrary{tikzmark}

\usepackage{stmaryrd}
\usepackage{bbm}
\usepackage{wrapfig}
\usepackage{pifont}

\definecolor{deepblue}{rgb}{0,0,0.5}
\definecolor{officeblue}{RGB}{0,102,204}
\definecolor{deepred}{rgb}{0.6,0,0}
\definecolor{deepgreen}{rgb}{0,0.5,0}
\definecolor{mybrickred}{RGB}{182,50,28}

\definecolor{fillcolor}{RGB}{216,217,252}

\usepackage{amsmath,amsfonts,bm}

\def\eqref#1{equation~\ref{#1}}

\def\1{\bm{1}}

\DeclareMathAlphabet{\mathsfit}{\encodingdefault}{\sfdefault}{m}{sl}
\SetMathAlphabet{\mathsfit}{bold}{\encodingdefault}{\sfdefault}{bx}{n}

\usepackage{pifont}

\title{Why Can GPT Learn In-Context? \\ Language Models Implicitly Perform Gradient Descent as Meta-Optimizers}

\author{Damai Dai$^\dag$\thanks{~~Contribution during internship at Microsoft Research.},~~Yutao Sun$^\parallel$\footnotemark[1],~~Li Dong$^\ddag$,~~Yaru Hao$^\ddag$,~~Shuming Ma$^\ddag$,~~Zhifang Sui$^\dag$,~~Furu Wei$^\ddag$ \\
$^\dag$ MOE Key Lab of Computational Linguistics, Peking University \\ 
$^\parallel$ Tsinghua University ~~~~ $^\ddag$ Microsoft Research \\
\texttt{\{daidamai,szf\}@pku.edu.cn} \\
\texttt{\{lidong1,fuwei\}@microsoft.com} \\}

\date{}

\begin{document}

\maketitle

\begin{abstract}

Large pretrained language models have shown surprising in-context learning (ICL) ability. 
With a few demonstration input-label pairs, they can predict the label for an unseen input without parameter updates. 
Despite the great success in performance, its working mechanism still remains an open question. 
In this paper, we explain language models as meta-optimizers and understand in-context learning as implicit finetuning. 
Theoretically, we figure out that Transformer attention has a dual form of gradient descent. 
On top of it, we understand ICL as follows: GPT first produces meta-gradients according to the demonstration examples, and then these meta-gradients are applied to the original GPT to build an ICL model. 
We comprehensively compare the behaviors of in-context learning and explicit finetuning on real tasks to provide empirical evidence that supports our understanding. 
Experimental results show that in-context learning behaves similarly to explicit finetuning from multiple perspectives. 
Inspired by the dual form between Transformer attention and gradient descent, we design a momentum-based attention by analogy with gradient descent with momentum. 
The improved performance over vanilla attention further supports our understanding from another perspective, and more importantly, shows the potential to utilize our understanding for future model design.
The code is available at \url{https://aka.ms/icl}. 

\end{abstract}

\section{Introduction}

\begin{figure}[t]
\centering
\includegraphics[width=0.98\linewidth]{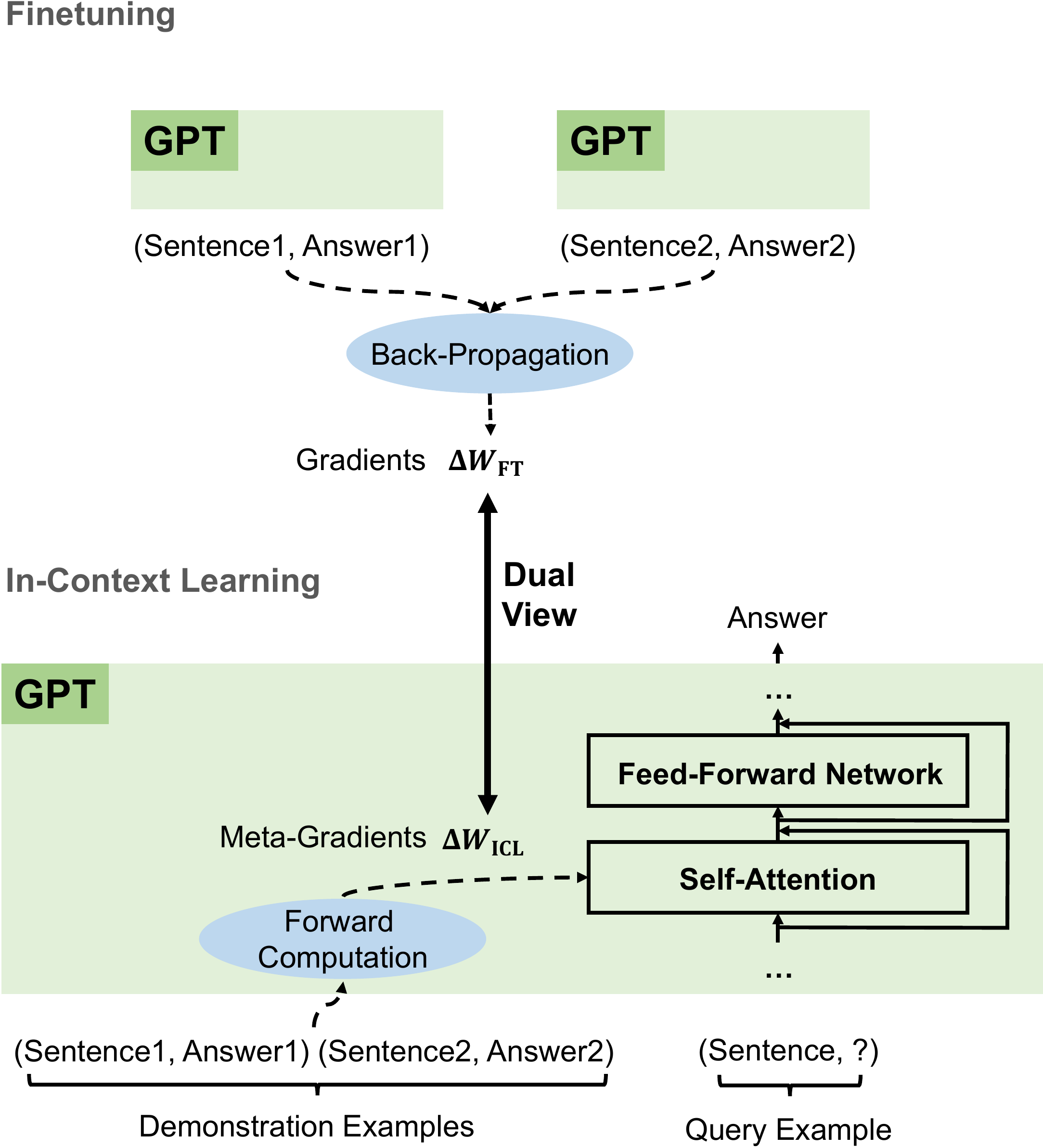}
\caption{
According to the demonstration examples, GPT produces meta-gradients for in-context learning (ICL) through forward computation. 
ICL works by applying these meta-gradients to the model through attention.
The meta-optimization process of ICL shares a dual view with finetuning that explicitly updates the model parameters with back-propagated gradients. 
}
\label{fig:dual_form}
\end{figure}

In recent years, large pretrained language models, especially in Transformer-based architectures (e.g., GPT; \citealt{gpt3}), have shown strong emergent in-context learning~(ICL) ability~\citep{wei2022emergent,icl_survey}. 
Different from finetuning which needs additional parameter updates, ICL just needs several demonstration examples prepended before the query input, and then the model can predict labels for unseen inputs. 
On numerous downstream tasks, large GPT models can achieve surprising performance, which even exceeds smaller models with supervised finetuning. 
However, although ICL has achieved great performance, its working mechanism is still an open question to be investigated. 

In this paper, we explain in-context learning as a process of meta-optimization and analyze connections between GPT-based in-context learning and finetuning. 
Concentrating on the attention modules, we figure out that the Transformer attention has a dual form of gradient descent. 
On top of it, we propose a novel perspective to explain in-context learning: 
(1) a pretrained GPT serves as a meta-optimizer; 
(2) it produces meta-gradients according to the demonstration examples through forward computation;  
(3) the meta-gradients are applied to the original language model through attention to build an ICL model. 
As illustrated in Figure~\ref{fig:dual_form}, in-context learning and explicit finetuning share a dual view of gradient descent, where ICL produces meta-gradients through forward computation, while finetuning computes gradients by back-propagation. 
Therefore, it is reasonable to understand in-context learning as implicit finetuning. 

In order to provide empirical evidence to support our understanding, we conduct comprehensive experiments based on real tasks.
On six classification tasks, we compare the model predictions, attention outputs, attention weights to query tokens, and attention weights to training tokens between in-context learning and finetuning. 
Experimental results validate that the behavior of in-context learning is similar to explicit finetuning from multiple perspectives. 
These results are strong evidence to prove the reasonability of our understanding of in-context learning as implicit finetuning. 

Further, inspired by the dual form between Transformer attention and gradient descent, we design a momentum-based attention, which regards the attention values as meta-gradients and applies the momentum mechanism~\citep{momentum, nesterov} to them. 
Experiments on both language modeling and in-context learning show that our momentum-based attention consistently outperforms vanilla attention, which supports our understanding of meta-optimization again from another perspective. 
We note that beyond this preliminary attempt, our understanding may have more potential to enlighten model design, which is worth investigating in the future. 

Our contributions are summarized as follows:
\begin{itemize}
\item We figure out a dual form between Transformer attention and gradient descent, and explain ICL as a process of meta-optimization. 
\item We analyze connections between in-context learning and explicit finetuning and propose to understand ICL as implicit finetuning. 
\item We provide several lines of empirical evidence to prove that ICL and explicit finetuning behave similarly from multiple perspectives. 
\item We design a momentum-based attention and validate its effectiveness, which supports our understanding of meta-optimization again and shows the potential of our understanding to enlighten future model design. 
\end{itemize}

\section{Background}

\subsection{In-Context Learning with GPT}
\label{sec:bg_icl}

In this paper, we focus on ICL for classification tasks using GPT~\citep{gpt3}. 
A GPT model is stacked with $L$ identical Transformer~\citep{transformer} decoder layers where each layer consists of an attention module and a feed-forward network. 
For a classification task, given a query input text $x$ and a candidate answer set $Y = \{ y_1, y_2, \dots, y_m \}$, we need to predict a label $\hat{y}$ conditional on $n$ demonstration examples $C = \{(x_1^{\prime}, y_1^{\prime}), (x_2^{\prime}, y_2^{\prime}), \dots, (x_n^{\prime}, y_n^{\prime})\}$, where $(x_i^{\prime}, y_i^{\prime})$ is an input-label pair different from the query one. 
Formally, given a GPT model $\mathcal{M}$, we first compute the probability of each answer $y_j$:
\begin{equation}
    P_{\mathcal{M}}(y_j \mid C, x).
\end{equation}
Since the label space is restricted for classification, we predict the final answer $\hat{y}$ by selecting the answer with the highest probability from the candidate answer set $Y$: 
\begin{equation}
    \hat{y} = \arg \max_{y_j} P_{\mathcal{M}}(y_j \mid C, x).
\end{equation}
In practice, we usually use a pre-defined template to format the demonstrations and prepend them before the query input. 
Let $\mathcal{T}(\cdot)$ be the function that formats an example, e.g.:
\begin{equation}
    \mathcal{T}(x, y) = \text{Sentence: } x. \; \text{Sentiment: } y.
\end{equation}
The contextual model input $I$ is organized like
\begin{equation}
    \mathcal{T}(x_1^{\prime}, y_1^{\prime}) \; \mathcal{T}(x_2^{\prime}, y_2^{\prime}) \; ... \; \mathcal{T}(x_n^{\prime}, y_n^{\prime}) \; \mathcal{T}(x, \_).
\end{equation}
Feeding this contextual input into $\mathcal{M}$, the probability of an answer $y_j$ is computed as
\begin{align}
    l_j = \mathcal{M} & (I) \cdot \mathbf{e}_{y_j}, \\
    P_{\mathcal{M}}(y_j \mid C, x) & = \operatorname{softmax}(l_j),
\end{align}
where $\mathcal{M}(I)$ denotes the output hidden state at the last token position; 
$\mathbf{e}_{y_j}$ denotes the output word embedding of $y_j$; 
and $l_j$ is the logit corresponding to the $j$-th answer. 

\subsection{Dual Form Between Attention and Linear Layers Optimized by Gradient Descent}
\label{sec:bg_dual}

The idea in this paper to explain language models as meta-optimizers is inspired by \citet{dual1964,dual_form}.
They present that linear layers optimized by gradient descent have a dual form of linear attention. 
Let $W_{0}, \Delta W \in \mathbb{R}^{d_{\text{out}} \times d_{\text{in}}}$ be the initialized parameter matrix and the update matrix, respectively, and $\mathbf{x} \in \mathbb{R}^{d_{\text{in}}}$ be the input representation. 
A linear layer optimized by gradient descent can be formulated as 
\begin{equation}
    \small
    \mathcal{F}(\mathbf{x}) = \left( W_{0} + \Delta W \right) \mathbf{x}. 
    \label{equ:dual_comp_1}
\end{equation}
In the back-propagation algorithm, $\Delta W$ is computed by accumulating the outer products of historic input representations $\mathbf{x}^{\prime T}_i \in \mathbb{R}^{d_{\text{in}}}$ and the error signals $\mathbf{e}_i \in \mathbb{R}^{d_{\text{out}}}$ of their corresponding outputs: 
\begin{equation}
    \small
    \Delta W = \sum_i \mathbf{e}_i \otimes \mathbf{x}^{\prime}_i, 
    \label{equ:dual_comp_2}
\end{equation}
where $\mathbf{e}_i$ is derived from the historic output gradients by multiplying $-\gamma$, the negative learning rate. 
Combing Equation~(\ref{equ:dual_comp_1}) and Equation~(\ref{equ:dual_comp_2}), we can derive the dual form of linear layers optimized by gradient descent: 
\begin{equation}
    \small
    \begin{aligned}
        \mathcal{F}(\mathbf{x}) = & \left( W_{0} + \Delta W \right) \mathbf{x} \\
        = & W_{0} \mathbf{x} + \Delta W \mathbf{x} \\
        = & W_{0} \mathbf{x} + \sum_i \left( \mathbf{e}_i \otimes \mathbf{x}^{\prime}_i\right) \mathbf{x} \\
        = & W_{0} \mathbf{x} + \sum_i \mathbf{e}_i \left( \mathbf{x}^{\prime T}_i \mathbf{x} \right) \\
        = & W_{0} \mathbf{x} + \operatorname{LinearAttn} \left( E, X^{\prime}, \mathbf{x} \right), 
    \end{aligned}
    \label{equ:sgd_attn_dual}
\end{equation}
where $\operatorname{LinearAttn}(V, K, \mathbf{q})$ denotes the linear attention operation, in which we regard the historic output error signals $E$ as values, the historic inputs $X^{\prime}$ as keys, and the current input $\mathbf{x}$ as the query. 

\section{Understanding In-Context Learning (ICL) as Implicit Finetuning}

We first qualitatively analyze the Transformer attention under a relaxed linear attention form to figure out a dual form between it and gradient descent.  
Then, we compare in-context learning with explicit finetuning to analyze connections between these two optimization forms. 
Based on these theoretical findings, we propose to understand in-context learning as implicit finetuning. 

\subsection{Understanding Transformer Attention as Meta-Optimization}
\label{sec:icl_dual}

Let $\mathbf{x} \in \mathbb{R}^{d}$ be the input representation of a query token $t$, and $\mathbf{q} = W_{Q} \mathbf{x} \in \mathbb{R}^{d^{\prime}}$ be the attention query vector. 
In the ICL setting, the attention result of a head is formulated as
\begin{equation}
    \small
    \begin{aligned}
        \mathcal{F}_{\text{ICL}}(\mathbf{q}) & = \operatorname{Attn}(V, K, \mathbf{q}) \\
        = & W_{V} [X^{\prime}; X] \operatorname{softmax} \left( \frac{\left( W_{K} [X^{\prime}; X] \right)^T \mathbf{q}}{\sqrt{d}} \right) ,
    \end{aligned}
\end{equation}
where $W_{Q}, W_{K}, W_{V} \in \mathbb{R}^{d^{\prime} \times d}$ are the projection matrices for computing the attention queries, keys, and values, respectively; 
$\sqrt{d}$ denotes the scaling factor; 
$X$ denotes the input representations of query tokens before $t$; 
$X^{\prime}$ denotes the input representations of the demonstration tokens; 
and $[X^{\prime}; X]$ denotes the matrix concatenation. 
For ease of qualitative analysis, we approximate the standard attention to relaxed linear attention by removing the softmax operation and the scaling factor:
\begin{equation}
    \small
    \begin{aligned}
    \mathcal{F}_{\text{ICL}}(\mathbf{q}) & \approx W_{V} [X^{\prime}; X] \left( W_{K} [X^{\prime}; X] \right)^T \mathbf{q} \\
    & = W_{V} X \left( W_{K} X \right)^T \mathbf{q} + W_{V} X^{\prime} \left( W_{K} X^{\prime} \right)^T \mathbf{q} \\
    & = \widetilde{\mathcal{F}}_{\text{ICL}}(\mathbf{q}).
    \end{aligned}
\end{equation}

We define $W_{\text{ZSL}} = W_{V} X \left( W_{K} X \right)^T$ as the initialized parameters to be updated since $W_{\text{ZSL}} \mathbf{q}$ is the attention result in the zero-shot learning~(ZSL) setting, where no demonstrations are given. 
Following the reverse direction of Equation~(\ref{equ:sgd_attn_dual}), we derive a dual form of the Transformer attention: 
\begin{equation}
    \small
    \begin{aligned}
        \widetilde{\mathcal{F}}_{\text{ICL}}(\mathbf{q}) & = W_{\text{ZSL}} \mathbf{q} + W_{V} X^{\prime} \left( W_{K} X^{\prime} \right)^T \mathbf{q} \\
        = & W_{\text{ZSL}} \mathbf{q} + \operatorname{LinearAttn} \left( W_{V} X^{\prime}, W_{K} X^{\prime}, \mathbf{q} \right) \\
        = & W_{\text{ZSL}} \mathbf{q} + \sum_i W_{V} \textbf{x}^{\prime}_i \left( \left( W_{K} \textbf{x}^{\prime}_i \right)^T \mathbf{q} \right) \\
        = & W_{\text{ZSL}} \mathbf{q} + \sum_i \left( (W_{V} \textbf{x}^{\prime}_i) \otimes \left( W_{K} \textbf{x}^{\prime}_i \right) \right) \mathbf{q} \\
        = & W_{\text{ZSL}} \mathbf{q} + \Delta W_{\text{ICL}} \mathbf{q} \\
        = & \left( W_{\text{ZSL}} + \Delta W_{\text{ICL}} \right) \mathbf{q}. 
    \end{aligned}
    \label{equ:icl_opti_dual}
\end{equation}
As shown in the above equations, the attention to the demonstration tokens is equivalent to parameter updates $\Delta W_{\text{ICL}}$ that take effect on $W_{\text{ZSL}}$. 
In addition, by analogy with $E$ in Equation~(\ref{equ:sgd_attn_dual}), we regard $W_{V} X^{\prime}$ as meta-gradients, which are used to compute the update matrix $\Delta W_{\text{ICL}}$. 

In summary, we explain in-context learning as a process of meta-optimization: 
(1) a pretrained GPT model serves as a meta-optimizer; 
(2) it produces meta-gradients according to the demonstration examples through forward computation;  
(3) through attention, the meta-gradients are applied to the original language model to build an ICL model. 

\subsection{Comparing ICL with Finetuning}

Based on the above understanding of in-context learning, we further compare the meta-optimization of in-context learning with the explicit optimization of finetuning to analyze connections between them. 
Considering that ICL directly takes effect on only the attention keys and values, we design a specific finetuning setting as the compared baseline, which also updates only the parameters for the key and value projection. 
Also in the relaxed linear attention form, the attention result of a finetuned head is formulated as
\begin{equation}
    \small
    \begin{aligned}
    \widetilde{\mathcal{F}}_{\text{FT}} (\mathbf{q}) & =
    (W_{V} + \Delta W_{V}) X X ^T (W_{K} + \Delta W_{K})^T \mathbf{q} \\
    & = \left( W_{\text{ZSL}} + \Delta W_{\text{FT}} \right) \mathbf{q} ,
    \end{aligned}
    \label{equ:ft}
\end{equation}
where $\Delta W_{K}$ and $\Delta W_{V}$ denote the parameter updates to $W_{K}$ and $W_{V}$, respectively, which are acquired by back-propagation from task-specific training objectives; and $\Delta W_{\text{FT}}$ is the updates to $W_{\text{ZSL}}$ introduced by finetuning.

For a more fair comparison with in-context learning, we further restrict the finetuning setting as follows: 
(1) we specify the training examples as the demonstration examples for in-context learning; 
(2) we train each example for only one step in the same order as demonstrated for in-context learning; 
(3) we format each training example with the same template used for ICL $\mathcal{T}(x^{\prime}_i, y^{\prime}_i)$ and use the causal language modeling objective for finetuning. 

Comparing in-context learning and this finetuning setting, we find that ICL has many properties in common with finetuning. 
We organize these common properties into the following four aspects.

\paragraph{Both Perform Gradient Descent}
Comparing Equation~(\ref{equ:icl_opti_dual}) and Equation~(\ref{equ:ft}), we find that both in-context learning and finetuning introduce updates ($\Delta W_{\text{ICL}}$ v.s. $\Delta W_{\text{FT}}$) to $W_{\text{ZSL}}$, which drive from implicit and explicit gradient descent, respectively. 
The main difference is that ICL produces meta-gradients by forward computation while finetuning acquires real gradients by back-propagation. 

\paragraph{Same Training Information}
The meta-gradients of ICL are produced according to the demonstration examples. 
The gradients of finetuning are also derived from the same training examples. 
That is to say, in-context learning and finetuning share the same source of training information. 

\paragraph{Same Causal Order of Training Examples}
In-context learning and our finetuning setting share the same causal order of training examples. 
ICL uses decoder-only Transformers so the subsequent tokens in the demonstrations will not affect the preceding ones. 
For our finetuning setting, we use the same order of training examples and train only one epoch, so we can also guarantee that the subsequent examples have no effect on the preceding ones. 

\paragraph{Both Aim at Attention}
Compared with zero-shot learning, the direct effect of in-context learning and our finetuning are both restricted to the computation of attention keys and values.  
For ICL, the model parameters are unchanged and it encodes demonstration information into additional keys and values to change the attention behavior. 
For finetuning, due to our restriction, the training information can be introduced to only the projection matrices for attention keys and values as well. 

Considering the above common properties between in-context learning and finetuning, we show that it is reasonable to understand in-context learning as implicit finetuning. 
In the rest of this paper, we compare ICL and explicit finetuning empirically from multiple perspectives to provide quantitative results to support this understanding. 

\section{Experiments}

\begin{table*}[t]
\centering
\setlength{\tabcolsep}{13pt}
\begin{tabular}{l | c c c c c c}
\toprule
& \textbf{SST2} & \textbf{SST5} & \textbf{MR} & \textbf{Subj} & \textbf{AGNews} & \textbf{CB} \\ 
\midrule
\# Validation Examples & 872 & 1101 & 1066 & 2000 & 7600 & 56 \\
\# Label Types & 2 & 5 & 2 & 2 & 4 & 3 \\
\midrule
ZSL Accuracy (GPT 1.3B) & 70.5 & 39.3 & 65.9 & 72.6 & 46.3 & 37.5 \\
FT Accuracy (GPT 1.3B) & 73.9 & 39.5 & 73.0 & 77.8 & 65.3 & 55.4 \\
ICL Accuracy (GPT 1.3B) & 92.7 & 45.0 & 89.0 & 90.0 & 79.2 & 57.1 \\
\midrule
ZSL Accuracy (GPT 2.7B) & 71.4 & 35.9 & 60.9 & 75.2 & 39.8 & 42.9 \\
FT Accuracy (GPT 2.7B) & 76.9 & 39.1 & 80.0 & 86.1 & 65.7 & 57.1 \\
ICL Accuracy (GPT 2.7B) & 95.0 & 46.5 & 91.3 & 90.3 & 80.3 & 55.4 \\
\bottomrule
\end{tabular}
\caption{
Statistics of six classification datasets (rows 1-2) and
validation accuracy in the zero-shot learning (ZSL), finetuning (FT), and in-context learning (ICL) settings on these datasets (rows 3-8).
}
\label{tab:dataset_acc}
\end{table*}

\begin{table*}[t]
\centering
\setlength{\tabcolsep}{12pt}
\begin{tabular}{l | c c c c c c c}
\toprule
\textbf{Model} & \textbf{SST2} & \textbf{SST5} & \textbf{MR} & \textbf{Subj} & \textbf{AGNews} & \textbf{CB} & \textbf{Average} \\ 
\midrule
GPT 1.3B & 91.84 & 66.67 & 97.08 & 87.17 & 83.08 & ~~87.50 & 85.56 \\
GPT 2.7B & 96.83 & 71.60 & 95.83 & 87.63 & 84.44 & 100.00 & 89.39 \\
\bottomrule
\end{tabular}
\caption{
Rec2FTP for two GPT models on six datasets. 
From the perspective of model prediction, ICL can cover most of the correct behavior of finetuning. 
}
\label{tab:rec2ftp}
\end{table*}

\subsection{Experimental Settings}

We analyze two off-the-shelf pretrained GPT models with 1.3 billion and 2.7 billion model parameters, respectively, which are released by fairseq\footnote{\url{https://github.com/facebookresearch/fairseq}}. 
In the rest of this paper, we call them GPT 1.3B and GPT 2.7B for short. 
All experiments are conducted on NVIDIA V100 GPUs with 32 GB memory. 

For each task, we use the same template to format examples for zero-shot learning~(ZSL), finetuning~(FT), and in-context learning~(ICL). 
Details of the templates used for each task are provided in Appendix~\ref{appendix:template}. 
The answer prediction processes for ZSL and finetuning are the same with ICL as described in Section~\ref{sec:bg_icl}, except that they do not have demonstration examples. 

For in-context learning, we fix the max number of demonstration examples to 32 and tune the random seed for each task to find a set of demonstration examples that achieves the best validation performance. 
For explicit finetuning, we use the same demonstration examples for in-context learning as the training examples and use SGD as the optimizer. 
For a fair comparison, we fine-tune the model for only one epoch and the 
training examples are provided in the same order as demonstrated for in-context learning. 
We tune the learning rate for finetuning and select the one that achieves the best validation performance. 
Details of the search range and selected value for the random seeds and learning rates are shown in Appendix~\ref{appendix:hyper}. 

\subsection{Evaluation Datasets}

We compare in-context learning and finetuning based on six datasets spanning three sorts of classification tasks. 
\textbf{SST2}~\citep{sst}, \textbf{SST5}~\citep{sst}, \textbf{MR}~\citep{mr} and \textbf{Subj}~\citep{subj} are four datasets for sentiment classification; 
\textbf{AGNews}~\citep{dbpedia:agnews} is a topic classification dataset; 
and \textbf{CB}~\citep{cb} is used for natural language inference. 
Statistics of the number of validation examples and label types are summarized in Table~\ref{tab:dataset_acc}.

For reference, we present the validation accuracy in the ZSL, finetuning, and ICL settings on six classification datasets in Table~\ref{tab:dataset_acc}. 
Compared with ZSL, ICL and finetuning both achieve considerable improvements, which means the optimizations they make are both helpful to these downstream tasks. 

\begin{table*}[t]
\centering
\setlength{\tabcolsep}{5.5pt}
\begin{tabular}{l l | c c c c c c c}
\toprule
\textbf{Model} & \textbf{Metric} & \textbf{SST2} & \textbf{SST5} & \textbf{MR} & \textbf{Subj} & \textbf{AGNews} & \textbf{CB} & \textbf{Average} \\ 
\midrule
\multirow{2}{*}{GPT 1.3B} & SimAOU (Random $\Delta$) & 0.002 & 0.003 & 0.001 & 0.002 & 0.002 & 0.003 & 0.002 \\
 & SimAOU ($\Delta$FT) & \textbf{0.110} & \textbf{0.080} & \textbf{0.222} & \textbf{0.191} & \textbf{0.281} & \textbf{0.234} & \textbf{0.186} \\
\midrule
\multirow{2}{*}{GPT 2.7B} & SimAOU (Random $\Delta$) & 0.000 & -0.002 & 0.000 & 0.001 & -0.002 & 0.000 & -0.001 \\
 & SimAOU ($\Delta$FT) & \textbf{0.195} & \textbf{0.323} & \textbf{0.157} & \textbf{0.212} & \textbf{0.333} & \textbf{0.130} & \textbf{0.225} \\
\bottomrule
\end{tabular}
\caption{
SimAOU for two GPT models on six datasets. 
ICL updates are much more similar to finetuning updates than to random updates. 
From the perspective of representation, ICL tends to change attention output representations in the same direction as finetuning changes. 
}
\label{tab:simaou}
\end{table*}

\begin{table*}[t]
\centering
\setlength{\tabcolsep}{4.3pt}
\begin{tabular}{l l | c c c c c c c}
\toprule
\textbf{Model} & \textbf{Metric} & \textbf{SST2} & \textbf{SST5} & \textbf{MR} & \textbf{Subj} & \textbf{AGNews} & \textbf{CB} & \textbf{Average} \\ 
\midrule
\multirow{2}{*}{GPT 1.3B} & SimAM (Before Finetuning) & 0.555 & 0.391 & 0.398 & 0.378 & 0.152 & 0.152 & 0.338 \\
 & SimAM (After Finetuning) & \textbf{0.585} & \textbf{0.404} & \textbf{0.498} & \textbf{0.490} & \textbf{0.496} & \textbf{0.177} & \textbf{0.442} \\
\midrule
\multirow{2}{*}{GPT 2.7B} & SimAM (Before Finetuning) & \textbf{0.687} & 0.380 & 0.314 & 0.346 & 0.172 & \textbf{0.228} & 0.355 \\
 & SimAM (After Finetuning) & \textbf{0.687} & \textbf{0.492} & \textbf{0.347} & \textbf{0.374} & \textbf{0.485} & 0.217 & \textbf{0.434} \\
\bottomrule
\end{tabular}
\caption{
SimAM for two models on six datasets. 
From the perspective of attention behavior, compared with attention weights before finetuning, ICL is more inclined to generate similar attention weights to those after finetuning. 
}
\label{tab:simam}
\end{table*}

\subsection{ICL Covers Most of Correct Predictions of Finetuning}

We compute a \textbf{recall to finetuning prediction (Rec2FTP)} to measure ICL can cover how much behavior of finetuning from the perspective of the model prediction. 
We first count $N_{\text{FT} > \text{ZSL}}$, the number of query examples that finetuning can predict correctly but ZSL cannot. 
Then, among these examples, we count $N_{(\text{FT} > \text{ZSL}) \wedge (\text{ICL} > \text{ZSL})}$, the number that ICL can also predict correctly. 
Finally, we compute the Rec2FTP score as $\frac{N_{(\text{FT} > \text{ZSL}) \wedge (\text{ICL} > \text{ZSL})}}{N_{\text{FT} > \text{ZSL}}}$. 
A higher Rec2FTP score suggests that ICL covers more correct behavior of finetuning from the perspective of the model prediction. 

We show the Rec2FTP scores for two GPT models on six datasets in Table~\ref{tab:rec2ftp}. 
As shown in the table, on average, ICL can correctly predict more than 85\% of the examples that finetuning can correct from ZSL. 
These results indicate that from the perspective of model prediction, ICL can cover most of the correct behavior of finetuning. 

\subsection{ICL Tends to Change Attention Outputs in the Same Direction as Finetuning}

From the perspective of representation, we compute a \textbf{similarity of the attention output updates (SimAOU)} to measure the similarity between the updates that ICL and finetuning make. 
For a query example, let $\mathbf{h}_{\text{X}}^{(l)}$ denote the normalized output representation of the last token at the $l$-th attention layer in setting X. 
The updates of ICL and finetuning compared with ZSL are $\mathbf{h}_{\text{ICL}}^{(l)} - \mathbf{h}_{\text{ZSL}}^{(l)}$ and $\mathbf{h}_{\text{FT}}^{(l)} - \mathbf{h}_{\text{ZSL}}^{(l)}$, respectively.
We compute the cosine between these two updates to get \textbf{SimAOU ($\Delta$FT)} at the $l$-th layer. 
A higher SimAOU ($\Delta$FT) means ICL is more inclined to update the attention output in the same direction as finetuning. 
For comparison, we also compute a baseline metric called \textbf{SimAOU (Random $\Delta$)} that computes the similarity between ICL updates and randomly generated updates. 

We present the SimAOU scores averaged across examples and layers for two GPT models on six datasets in Table~\ref{tab:simaou}. 
From the table, we find that SimAOU (Random $\Delta$) is always around zero, while SimAOU ($\Delta$FT) remains much more positive. 
These results indicate that ICL updates are much more similar to finetuning updates than to random updates. 
From the perspective of representation, we prove that ICL tends to change the attention outputs in the same direction as finetuning. 

\begin{table*}[t]
\centering
\setlength{\tabcolsep}{5pt}
\begin{tabular}{l l | c c c c c c c}
\toprule
\textbf{Model} & \textbf{Metric} & \textbf{SST2} & \textbf{SST5} & \textbf{MR} & \textbf{Subj} & \textbf{AGNews} & \textbf{CB} & \textbf{Average} \\ 
\midrule
\multirow{2}{*}{GPT 1.3B} & Kendall (ICL, Random) & 0.000 & -0.001 & 0.000 & 0.001 & -0.001 & 0.000 & 0.000 \\
 & Kendall (ICL, FT) & \textbf{0.192} & \textbf{0.151} & \textbf{0.173} & \textbf{0.181} & \textbf{0.190} & \textbf{0.274} & \textbf{0.193} \\
\midrule
\multirow{2}{*}{GPT 2.7B} & Kendall (ICL, Random) & -0.001 & 0.000 & 0.000 & 0.000 & 0.000 & -0.001 & 0.000 \\
 & Kendall (ICL, FT) & \textbf{0.213} & \textbf{0.177} & \textbf{0.264} & \textbf{0.203} & \textbf{0.201} & \textbf{0.225} & \textbf{0.214} \\
\bottomrule
\end{tabular}
\caption{
Kendall rank correlation coefficients for two GPT models on six datasets. 
Compared with random attention weights, ICL attention weights to training tokens are much more similar to finetuning attention weights. 
}
\label{tab:training_attn}
\end{table*}

\subsection{ICL Is Inclined to Generate Similar Attention Weights to Finetuning}

From the perspective of attention behavior, we compute a \textbf{similarity of the attention map (SimAM)} to measure the similarity of the attention map to query tokens for ICL and finetuning. 
For a query example, let $\mathbf{m}_{\text{X}}^{(l, h)}$ denote the attention weights before softmax of the last token at the $h$-th attention head in the $l$-th attention layer in setting X. 
For ICL, we omit the attention to the demonstration tokens and only monitor the attention weights to the query tokens. 
First, before finetuning, we compute the cosine between $\mathbf{m}_{\text{ICL}}^{(l, h)}$ and $\mathbf{m}_{\text{ZSL}}^{(l, h)}$ and then average the similarity across attention heads to get \textbf{SimAM (Before Finetuning)} at each layer. 
Similarly, after finetuning, we compute the cosine between $\mathbf{m}_{\text{ICL}}^{(l, h)}$ and $\mathbf{m}_{\text{FT}}^{(l, h)}$ to get \textbf{SimAM (After Finetuning)}. 
A higher SimAM (After Finetuning) over SimAM (Before Finetuning) indicates that the attention behavior of ICL is more similar to a finetuned model than a non-finetuned one. 

Table~\ref{tab:simam} demonstrates the SimAM scores averaged across examples and layers for two GPT models on six datasets. 
We observe that compared with attention weights before finetuning, ICL is more inclined to generate similar attention weights to attention weights after finetuning. 
Again, from the perspective of attention behavior, we prove that ICL behaves similarly to finetuning. 

\subsection{ICL and Finetuning Tend to Pay Similar Attention to Training Tokens}

Since we understand ICL as a process of meta-optimization, we also compare the attention to training tokens for ICL and finetuning with the \textbf{Kendall rank correlation coefficient}~\citep{kendall}. 
For a query example, let $\mathbf{m}_{\text{ICL}}^{(l)}$ denote the ICL attention weights to the demonstration tokens of the last query token in the $l$-th attention layer, which is summed across attention heads. 
For finetuning, we first record all the attention queries ${Q^{\prime}}^{(l, h)} \in \mathbb{R}^{d^{\prime} \times N}$ of the training tokens, and then use the inner product between them and the attention query $\mathbf{q}^{(l, h)} \in \mathbb{R}^{d^{\prime}}$ of the last token in the query example as the finetuning attention weights to the training tokens: 
$\mathbf{m}_{\text{FT}}^{(l)} = \sum_{h}{{Q^{\prime}}^{(l, h)}}^T \mathbf{q}^{(l, h)}$, which is also summed across attention heads. 
The Kendall coefficient between $\mathbf{m}_{\text{ICL}}^{(l)}$ and $\mathbf{m}_{\text{FT}}^{(l)}$ is computed as $\textbf{Kendall (ICL, FT)}=\frac{P_c-P_d}{N(N-1)/2}$, where $N$ denotes the number of training tokens, $P_c$ denotes the number of concordant pairs, and $P_d$ denotes the number of discordant pairs. 
A higher Kendall coefficient means that the orders of attention weights to training tokens of ICL and finetuning are more similar.
For comparison, we also compute the Kendall coefficient between $\mathbf{m}_{\text{ICL}}^{(l)}$ and randomly generated attention weights $\mathbf{m}_{\text{Random}}^{(l)}$, which we call \textbf{$\text{Kendall (ICL, Random)}$}. 

Table~\ref{tab:training_attn} shows the Kendall correlation coefficients averaged across examples and layers for two GPT models on six datasets. 
We find that Kendall (ICL, Random) is always near zero, while Kendall (ICL, FT) always maintains a distinctly positive value. 
These results suggest that ICL and finetuning tend to pay similar attention to training tokens. 




\section{Momentum-Based Attention Inspired by Dual Form of Transformer Attention}

\begin{figure}[t]
\centering
\includegraphics[width=0.98\linewidth]{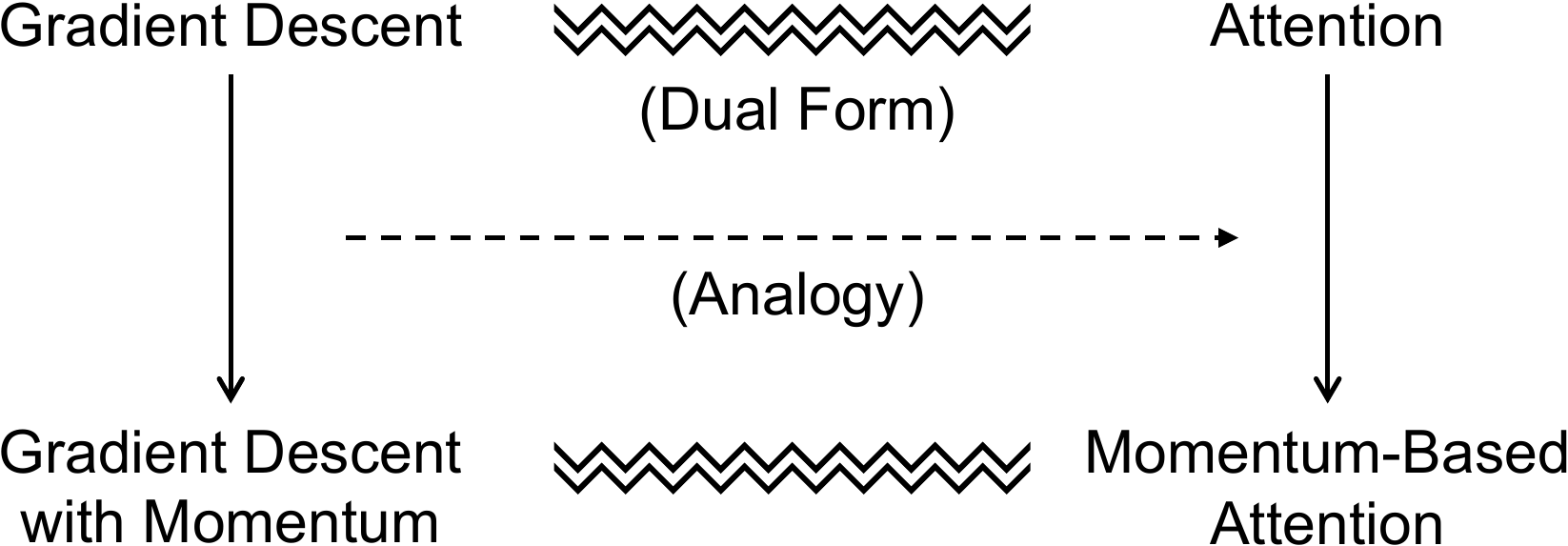}
\caption{
Inspired by the dual form between attention and gradient descent, we introduce the momentum mechanism into Transformer attention by analogy with gradient descent with momentum. 
}
\label{fig:momentum}
\end{figure}

We have figured out the dual form between Transformer attention and gradient descent. 
As illustrated in Figure~\ref{fig:momentum}, inspired by this dual view, we investigate whether we can utilize momentum~\citep{momentum, nesterov}, a widely used technique for optimization algorithms, to improve Transformer attention. 

Gradient descent with momentum averages gradients among timestamps: 
\begin{equation}
    \small
    \Theta_t = \Theta_{t-1} - \gamma \sum_{i=1}^{t-1} \eta^{t-i} \nabla f_{\Theta_{i}},
\end{equation}
where $\gamma$ is the learning rate and $\eta$ is a scalar between 0 and 1. 
As stated in Section~\ref{sec:icl_dual}, the attention values serve as meta-gradients. 
By analogy with gradient descent with momentum, we try to use Exponential Moving Average (EMA; \citealt{ema}) to average the attention values to build the momentum-based attention: 
\begin{equation}
\small
\begin{aligned}
\operatorname{MoAttn}(V, K, \mathbf{q}_t) & = \operatorname{Attn}(V, K, \mathbf{q}_t) + \operatorname{EMA}(V) \\ 
& = V \operatorname{softmax}(\frac{K^T \mathbf{q}_t}{\sqrt{d}}) + \sum_{i=1}^{t-1} \eta^{t-i} \mathbf{v}_i, \nonumber
\end{aligned}
\end{equation}
where $\mathbf{v}_i$ is the $i$-th attention value vector. 
The momentum of attention value vectors explicitly strengthens the recency bias of attention, which has been shown helpful for language modeling~\citep{alibi}. 
Therefore, we assume that introducing momentum into attention will contribute to faster convergence and better performance. 

\begin{table*}[t]
    \centering
    \setlength{\tabcolsep}{19pt}
    \begin{tabular}{c c c c c}
    \toprule
    \textbf{Model} & \textbf{Train$_{1024}$} & \textbf{Valid$_{256}$} & \textbf{Valid$_{512}$} & \textbf{Valid$_{1024}$} \\ 
    \midrule
    Transformer & 17.61 & 19.50 & 16.87 & 15.14 \\
    Transformer$_{\operatorname{MoAttn}}$ & \textbf{17.55} & \textbf{19.37} & \textbf{16.73} & \textbf{15.02} \\ 
    \bottomrule
    \end{tabular}
    \caption{
    Perplexity on the training set and validation sets with different input lengths for language modeling. 
    Momentum-based attention achieves a consistent perplexity improvement compared with the vanilla Transformer. 
    }
    \label{tab:lm_momentum}
\end{table*}

\begin{table*}[t]
    \centering
    \setlength{\tabcolsep}{10pt}
    \begin{tabular}{c c c c c c c c}
    \toprule
    \textbf{Model} & \textbf{SST5} & \textbf{IMDB} & \textbf{MR} & \textbf{CB} & \textbf{ARC-E} & \textbf{PIQA} & \textbf{Average} \\ 
    \midrule
    Transformer & 25.3 & 64.0 & 61.2 & 43.9 & 48.2 & 68.7 & 51.9 \\
    Transformer$_{\operatorname{MoAttn}}$ & \textbf{27.4} & \textbf{70.3} & \textbf{64.8} & \textbf{46.8} & \textbf{50.0} & \textbf{69.0} & \textbf{54.7} \\ 
    \bottomrule
    \end{tabular}
    \caption{
    Accuracy on six in-context learning datasets. 
    Introducing momentum into attention improves the accuracy of the vanilla Transformer by 2.8 on average. 
    }
    \label{tab:icl_momentum}
\end{table*}

\paragraph{Experiments on Language Modeling}
First, we evaluate the effect of momentum-based attention on language modeling. 
We train two GPT models with 350M parameters from scratch, where one is the vanilla Transformer, and another applies momentum to attention. 
More training details are provided in Appendix~\ref{appendix:scratch}. 
We evaluate the perplexity of these two models on the training set and three validation sets with input lengths of 256, 512, and 1024, respectively. 
The results are shown in Table~\ref{tab:lm_momentum}. 
On all of the validation sets, applying momentum to attention introduces a consistent perplexity improvement compared with the vanilla Transformer. 

\paragraph{Experiments on In-Context Learning}
We also evaluate the in-context learning ability of the above language models to verify the effectiveness of momentum-based attention on downstream tasks. 
We consider six datasets for sentiment analysis (SST5~\citep{sst}, IMDB~\citep{imdb}, and MR~\citep{mr}), natural language inference (CB~\citep{cb}), and multi-choice selection (ARC-E~\citep{arc} and PIQA~\citep{piqa}). 
For all of these datasets, we use up to 32 examples as demonstrations. 
As shown in Table~\ref{tab:icl_momentum}, compared with vanilla Transformer, using momentum-based attention achieves consistently higher accuracy on all of these datasets. 

The performance improvements on both language modeling and in-context learning prove our deduction that introducing momentum will improve Transformer attention. 
From another perspective, these results further support our understanding of Transformer attention as meta-optimization. 

\section{Related Work}

Recently, some pieces of work have attempted to understand the inference mechanism of in-context learning. 
\citet{bayesian} explain in-context learning as implicit Bayesian inference. 
They state that in-context learning emerges when language models can infer the shared latent concept among the demonstration examples, which is learned during pretraining. 
On another aspect, \citet{olsson2022induction} focus on specific modules in Transformers. 
They find some induction heads in Transformers that refer to abstract patterns in previous sequences to help predict the next token. 
They indicate that the induction heads drive the ability of in-context learning. 
Different from them, we concentrate on the learning algorithm of ICL and explain it as a process of meta-optimization. 

Some other work also studies the learning algorithm of ICL. 
As a case study, \citet{icl_case_study} show that Transformers can be trained to in-context learn a class of linear functions and the performance is comparable to the least squares estimator. 
Based on linear regression, \citet{icl_learning_algorithm} prove that they can construct parameters of Transformers to implement gradient-descent-based learning algorithms. 
Further, they show that models trained with an in-context learning objective tend to match the behavior of models computed by explicit learning algorithms. 
Also based on regression tasks, \citet{icl_gd} show that linear attention-only Transformers with constructed parameters that implement gradient descent and models learned by an in-context learning objective are highly related. 
Compared with them, we are the first ones to explain in-context learning in real scenarios. 
To be specific, (1) we analyze in-context learning for off-the-shelf GPT models, instead of models trained from scratch by an ICL objective; (2) our experiments are based on real NLP tasks, instead of toy ones like linear regression. 

\section{Conclusion}

In this paper, we aim to explain the working mechanism of GPT-based ICL.
Theoretically, we figure out a dual form between Transformer attention and gradient descent, and propose to understand ICL as a process of meta-optimization. 
Further, we analyze connections between ICL and explicit finetuning and show the reasonability to regard ICL as implicit finetuning. 
Empirically, we comprehensively compare ICL and finetuning based on six real NLP tasks. 
The results prove that ICL behaves similarly to explicit finetuning from multiple perspectives. 
Further, inspired by our understanding of meta-optimization, we design a momentum-based attention that achieves consistent performance improvements over vanilla attention. 
We believe our understanding will have more potential to enlighten ICL applications and model design in the future. 

\section*{Limitations}

Although the ability of in-context learning has been found for different architectures (e.g., Transformer and LSTM), we consider only Transformer-based in-context learning in this paper because Transformer is the current mainstream architecture of NLP. 
However, as for in-context learning itself, figuring out how it works for other architectures is also a meaningful problem, which we encourage to study in the future. 

As for the dual form we point out between Transformer attention and gradient descent, we consider a relaxed form of linear attention for qualitative analysis. 
Although the experimental results support our understanding well, the mechanism of standard Transformer attention without approximation may be more complex and should be studied more clearly in the future. 

As for empirical experiments, our analysis needs to record a large number of intermediate results (e.g., attention output representations, and attention weights to query tokens and demonstration tokens) for thousands of validation examples. 
Considering the storage space and computational cost of analysis, we only analyze GPT models with up to 2.7B parameters and leave larger models such as GPT 13B for future work. 
In addition, for the clarity of the problem definition and the convenience of experiments, our analysis is based on only classification tasks. 
Although classification is a representative application of in-context learning, other tasks like multiple choice and open-ended generation are not considered in this paper and could be investigated in the future. 

\section*{Acknowledgement}

Damai Dai and Zhifang Sui are supported by the National Key Research and Development Program of China 2020AAA0106700 and NSFC project U19A2065. 

\nocite{general_purpose}
\nocite{meta_learning_bp}

\bibliography{anthology,icl}
\bibliographystyle{acl_natbib}

\clearpage
\appendix

\section*{Appendix}

\section{Templates for In-Context Learning}
\label{appendix:template}

We demonstrate the templates used to format examples and the candidate answer sets for six classification datasets used in our experiments in Table~\ref{tab:template}. 

\begin{table*}[ht]
    \centering
    \begin{tabularx}{\textwidth}{l X X}
    \toprule
    \textbf{Dataset} & \textbf{Template} & \textbf{Candidate Answer Set} \\
    \midrule
    SST2 & Sentence: \{Sentence\} & \{ Negative, Positive \} \\
    &Label: \{Label\} & \\
    \midrule
    SST5 & Sentence: \{Sentence\} & \{ terrible, bad, neutral, good, great \} \\
    & Label: \{Label\} & \\
    \midrule
    MR & Review: \{Sentence\} & \{ Negative, Positive \} \\
    & Sentiment: \{Label\} & \\
    \midrule
    Subj & Input: \{Sentence\} & \{ objective, subjective \} \\
    & Type: \{Label\} & \\
    \midrule
    AGNews & Classify the news articles into the categories of World, Sports, Business, and Technology. & \{ World, Sports, Business, Technology \} \\
    & News: \{Sentence\} & \\
    & Type: \{Label\} & \\
    \midrule
    CB & \{Premise\} & \{ True, False, Neither \} \\
    & Question: \{Hypothesis\} True, False, or Neither? & \\
    & Answer: \{Label\} & \\
    \bottomrule
    \end{tabularx}
    \caption{Formatting templates and candidate answer sets for six classification datasets. }
    \label{tab:template}
\end{table*}

\section{Hyper-Parameters for In-Context Learning and Finetuning}
\label{appendix:hyper}

We perform grid search to find the best random seed for ICL and the best learning rate for finetuning. 
The search range for all the datasets is the same.
For random seeds, we search in $\{ 1, 2, 3, 4, 5, 6, 7 \}$. 
For learning rates, the search base values are $\{ 1, 2, 3, 4, 5, 6, 7, 8, 9 \}$ and we scale them to $0.1$, $0.01$, $0.001$, and $0.0001$ times, i.e., we have $9 \times 4 = 36$ values to search. 
As an exception, for GPT 1.3B finetuned on SST5, we perform a more fine-grained search and finally set its learning rate to 0.00016 since the finetuned model cannot outperform the zero-shot learning with the above 36 learning rates. 

In Table~\ref{tab:hyper}, we present the details of the selected random seeds and learning rates for two GPT models on six classification datasets.   

\begin{table*}[ht]
    \centering
    \begin{tabular}{l l c c}
    \toprule
    \textbf{Hyper-Parameter} & \textbf{Dataset} & \textbf{GPT 1.3B} & \textbf{GPT 2.7B} \\
    \midrule
    \multirow{6}{*}{Random Seed} & SST2 & 2 & 7 \\
     & SST5 & 5 & 5 \\
     & MR & 5 & 1 \\
     & Subj & 4 & 4 \\
     & AGNews & 3 & 3 \\
     & CB & 3 & 3 \\
    \midrule
    \multirow{6}{*}{Learning Rate} & SST2 & 0.0005 & 0.007 \\
     & SST5 & ~~0.00016 & 0.04~~ \\
     & MR & 0.003~~ & 0.001 \\
     & Subj & 0.003~~ & 0.002 \\
     & AGNews & 0.2~~~~~~ & 0.2~~~~ \\
     & CB & 0.08~~~~ & 0.01~~ \\
    \bottomrule
    \end{tabular}
    \caption{
    Selected random seeds and learning rates for two GPT models on six classification datasets. 
    }
    \label{tab:hyper}
\end{table*}

\section{Hyper-Parameters for Training Language Models from Scratch}
\label{appendix:scratch}

The hyper-parameters for training two language models from scratch are summarized in Table~\ref{tab:hyper_scratch}. 

\begin{table*}[ht]
    \centering
    \setlength{\tabcolsep}{10pt}
    \begin{tabular}{l c}
    \toprule
    \textbf{Hyper-parameter} & \textbf{Value} \\
    \midrule
    Embedding \& Hidden Dimension & 1024 \\
    FFN Inner Hidden Dimension & 4096 \\
    Number of Attention Heads & 16 \\
    Number of Transformer Layers & 24 \\
    Number of Parameters & 350M \\
    \midrule
    Sequence Length & 1024 \\
    Batch Size & 512K Tokens \\
    \midrule
    Optimizer & Adam \\
    Adam Betas & (0.9, 0.98) \\
    Adam Epsilon & 1e-6 \\
    Maximum Learning Rate & 3e-4 \\
    Learning Rate Scheduler & Polynomial Decay \\
    Total Training Steps & 500K \\
    Warm-up Steps & 20K \\
    Gradient Clip Norm & 2.0 \\
    \bottomrule
    \end{tabular}
    \caption{
    Hyper-parameters for training two language models from scratch. 
    }
    \label{tab:hyper_scratch}
\end{table*}

\end{document}